\documentclass[10pt,twocolumn,letterpaper]{article}

\usepackage{iccv}      % To produce the REVIEW version
\usepackage{amsmath}
\usepackage{xcolor}
\usepackage{listings}
\usepackage{multirow}
\usepackage[utf8]{inputenc}
\usepackage{tcolorbox}
\usepackage{xcolor}
\usepackage{booktabs}
\definecolor{iccvblue}{rgb}{0.21,0.49,0.74}
\usepackage[pagebackref,breaklinks,colorlinks,allcolors=iccvblue]{hyperref}

\def\paperID{} % *** Enter the Paper ID here
\def\confName{ICCV}
\def\confYear{2025}

\title{PHT-CAD: Efficient CAD Parametric Primitive Analysis with Progressive Hierarchical Tuning}
\author{Ke Niu\thanks{Equal Contribution \newline \hspace*{0.35cm}\dag Corresponding Author}\ , Yuwen Chen$^{*}$, Haiyang Yu, Zhuofan Chen, Xianghui Que, Bin Li, Xiangyang Xue$^{\dag}$\\
Shanghai Key Laboratory of Intelligent Information Processing\\ School of Computer Science, Fudan University\\
{\tt\small \{kniu22, ywchen23, zfchen23, xhque23\}@m.fudan.edu.cn}\\
{\tt\small \{hyyu20, libin, xyxue\}@fudan.edu.cn}}

\begin{document}
\maketitle

\begin{abstract}
Computer-Aided Design (CAD) plays a pivotal role in industrial manufacturing, yet 2D Parametric Primitive Analysis (PPA) remains underexplored due to two key challenges: structural constraint reasoning and advanced semantic understanding. To tackle these challenges, we first propose an Efficient Hybrid Parametrization (EHP) for better representing 2D engineering drawings. EHP contains four types of atomic component (\textit{i.e.}, point, line, circle, and arc). Additionally, we propose PHT-CAD, a novel 2D PPA framework that harnesses the modality alignment and reasoning capabilities of Vision-Language Models (VLMs) for precise engineering drawing analysis. In PHT-CAD, we introduce four dedicated regression heads to predict corresponding atomic components. To train PHT-CAD, a three-stage training paradigm Progressive Hierarchical Tuning (PHT) is proposed to progressively enhance PHT-CAD's capability to perceive individual primitives, infer structural constraints, and align annotation layers with their corresponding geometric representations. Considering that existing datasets lack complete annotation layers and real-world engineering drawings, we introduce ParaCAD, the first large-scale benchmark that explicitly integrates both the geometric and annotation layers. ParaCAD comprises over 10 million annotated drawings for training and 3,000 real-world industrial drawings with complex topological structures and physical constraints for test. Extensive experiments demonstrate the effectiveness of PHT-CAD and highlight the practical significance of ParaCAD in advancing 2D PPA research. The code and adopted datasets are available at \url{https://github.com/yuwen-chen616/PHT-CAD}.

\end{abstract}
\section{Introduction}
\label{sec:intro}

Computer-Aided Design (CAD) has become an essential tool in modern industrial product design, playing a pivotal role across various industries. In particular, 2D engineering drawings serve as a fundamental component in industrial manufacturing, especially in fields such as industrial equipment production (\textit{e.g.}, laser cutting and PCB design). Moreover, quality control in industrial manufacturing heavily depends on the precise measurement and inspection of geometric dimensions and tolerances specified in 2D engineering drawings. With the rapid advancement of deep learning, numerous 2D CAD-related tasks have garnered significant research attention, including segmentation\cite{zhang2023component}, 2D sketch generation\cite{seff2021vitruvion}, and 2D-to-3D code generation~\cite{chen2024img2cad}. These advancements have significantly improved automation and efficiency in CAD-related applications, yet challenges remain in achieving accurate and reliable parametric analysis for real-world engineering drawings.

\begin{figure}[t]
    \centering
    \includegraphics[width=0.47\textwidth]{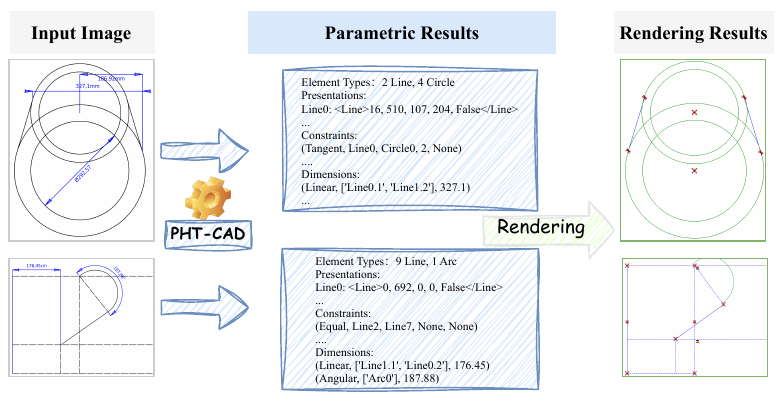} 
    \caption{Comparison between sketches/hand-drawn images and structured engineering drawings. }
    \label{fig:sketch_vs_drawing}
\end{figure}

Unlike the aforementioned tasks that primarily focus on CAD sketches, 2D Parametric Primitive Analysis (PPA)~\cite{wang2024parametric} is dedicated to processing 2D engineering drawings, which not only contain geometric primitives and constraints (\textit{i.e.}, the geometric layer, as found in CAD sketches) but also dimensional parameter annotations (\textit{i.e.}, the annotation layer) that are essential for real-world engineering applications. The challenges of 2D PPA mainly stem from two key aspects:
(1) \textit{Structural Constraint Reasoning}: Engineering drawings inherently encode complex inter-dependencies among primitives through geometric constraints, such as parallelism, tangency, and coincidence. 
(2) \textit{Advanced Semantic Understanding}: Beyond recognizing geometric primitives, PPA requires a deeper comprehension of dimensional annotations, implicit constraints, and hierarchical relationships between primitives.

Building on the modality alignment and semantic reasoning capabilities of Vision-Language Models (VLMs)~\cite{yu2025eve,niu2025synthesizing,fu2023denoising,fu2025foundation,niu2025chatreid,yu2025umit}, we introduce PHT-CAD, a novel 2D PPA paradigm designed to leverage VLMs for interactive engineering drawing analysis. The PHT-CAD framework adopts a standard VLM architecture, integrating a vision encoder and a VLM to facilitate multi-modal understanding of CAD data.
To achieve a more efficient representation of CAD parametric data while seamlessly aligning with the VLM paradigm, we introduce Efficient Hybrid Parametrization (EHP), a novel strategy for structured parametric encoding of 2D engineering drawings. EHP represents 2D engineering drawings by four types of atomic components: point, line, circle, and arc. Furthermore, recognizing the inherent limitations of VLMs in precise numerical prediction, we introduce four dedicated regression heads to predict corresponding atomic components and propose the Parametric Mean Squared Error (P-MSE) Loss to supervise numerical predictions.

For training PHT-CAD, we develop Progressive Hierarchical Tuning (PHT), which is designed to systematically enhance PHT-CAD’s capability in parametric primitive analysis. The training process is structured into three progressive stages: 1) \textit{Primitive Perception Tuning}, which enables PHT-CAD to perceive individual geometric primitives and their parameters; 2) \textit{Structural Perception Tuning}, where PHT-CAD is trained on unlabeled engineering drawings (\textit{i.e.}, sketches) to learn structural constraints between primitives; 3) \textit{Annotation-geometry Alignment}, which equips PHT-CAD with the ability to associate annotations layers with their corresponding geometric layers.
Through these three training stages, PHT-CAD's intrinsic geometric perception and reasoning capabilities are progressively enhanced, enabling fine-grained primitive parameterization.

Another challenge impeding the progress of PPA is the lack of open-source 2D engineering drawing datasets. Most existing approaches primarily focus on processing sketches or hand-drawn images, which significantly differ from engineering drawings. To address this gap, we introduce ParaCAD, the first large-scale benchmark for 2D PPA that explicitly incorporates the annotation layer. The raw data of the training set in ParaCAD are derived from two large-scale open-source datasets, SketchGraphs\cite{seff2020sketchgraphs} and CADL\cite{ganin2021computer}. We propose the DrawParam methodology to generate parameterization annotations for all samples of these two datasets, resulting in a training set containing over 10 million annotated engineering drawings. To ensure alignment with real-world industrial design applications, the training set of ParaCAD includes 5 million color-coded drawings with enhanced annotation visibility. Additionally, we collect a test set comprising 3,000 real-world industrial engineering drawings, featuring greater topological complexity than existing datasets and adhering to real-world design intent and physical constraints, such as fully enclosed structures.

The contributions of this paper can be summarized as follows:

\begin{itemize} 
    \item We introduce PHT-CAD, an innovative 2D parametric primitive analysis (PPA) paradigm that leverages Vision-Language Models (VLMs) for interactive engineering drawing analysis. 
    
    \item We propose a novel representation approach, termed Efficient Hybrid Parametrization (EHP), for 2D PPA. To generate more accurate numerical predictions in EHP, we introduce four dedicated regression heads and develop a Parametric Mean Squared Error loss to supervise them.

    \item We present ParaCAD, the first large-scale 2D PPA benchmark that incorporates the annotation layer. ParaCAD includes 10.26 million annotated engineering drawings for training and 3,000 real-world industrial drawings for test. The samples for test exhibit higher topological complexity and adhere to physical constraints.

\end{itemize} 

\begin{figure*}[t]
    \centering
    \includegraphics[width=0.8\linewidth]{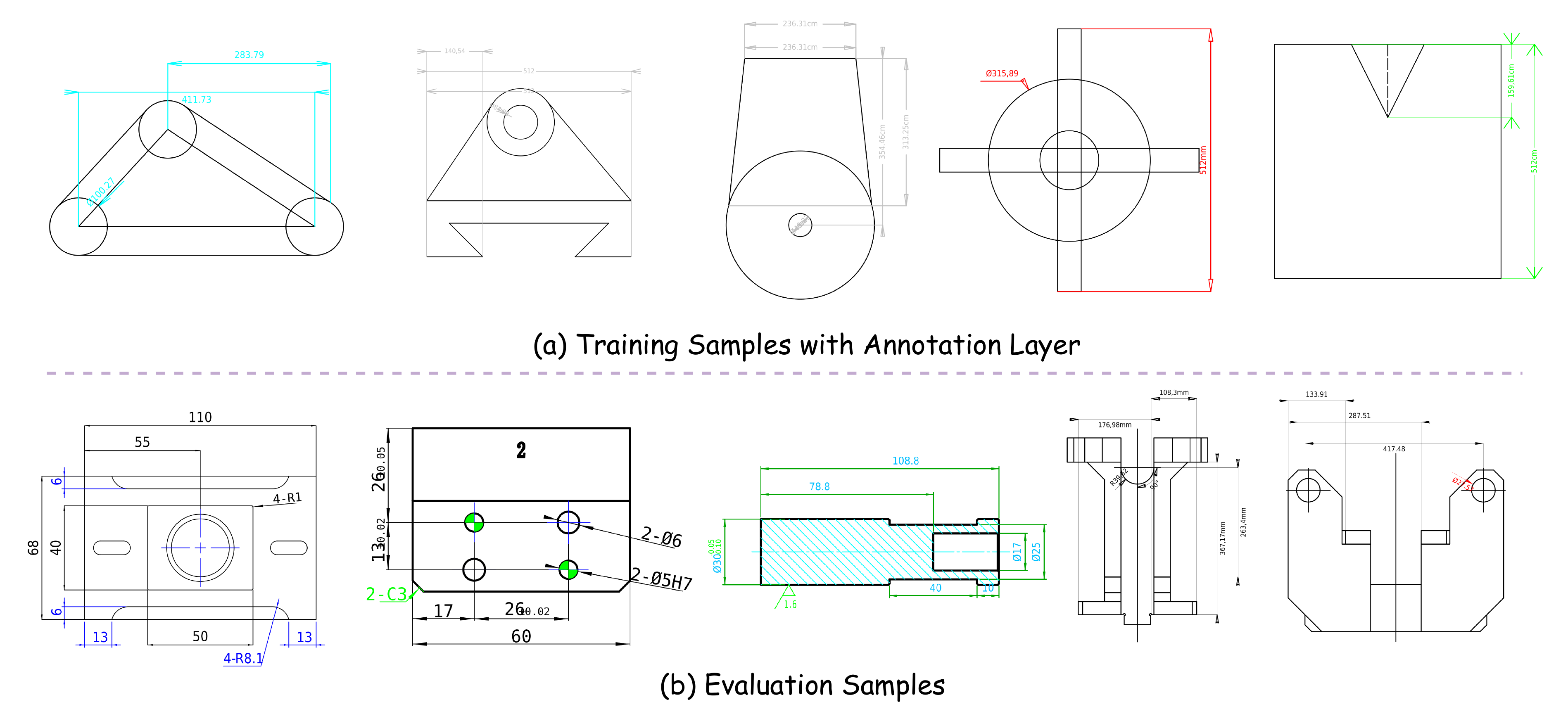}
\caption{Some samples of ParaCAD.
}
\label{sample}
\end{figure*}

\section{Related Work}
\label{sec:formatting}
\subsection{Applications of LVLMs in CAD-Related Tasks}
Recent advances in vision-language models (VLMs) have significantly impacted CAD-related research. 
Several existing methods have been specifically designed to address particular CAD-related tasks.
CAD2Program~\cite{wang20242d} generates 3D parametric models from 2D CAD drawings.
CAD-Talk~\cite{yuan2024cadtalk} segments CAD programs into code blocks corresponding to semantically meaningful shape parts. 
Img2CAD~\cite{chen2024img2cad} proposes a structured visual geometry approach to generate editable 3D CAD models from single images. 
LLM4CAD~\cite{li2025llm4cad} generates 3D CAD models from both textual descriptions and image inputs.
Some studies have begun exploring VLM-based approaches for a unified framework to tackle multiple CAD tasks.
CAD-MLLM~\cite{xu2024cad} enables parametric CAD modeling from text, images, and point clouds, and proposes the Omni-CAD dataset. 
CadVLM~\cite{wu2024cadvlm} extends CAD-MLLM by incorporating visual modalities to enhance functionalities such as CAD sketch auto-completion, automatic constraint application, and image-guided generation. 
CAD-Assistant~\cite{mallis2024cad} introduces a novel approach that leverages VLMs as planners, where the model interacts with Python APIs to accomplish various CAD tasks.

\subsection{2D CAD Sketch Parametric Primitive Analysis}
2D engineer drawings play a crucial role in the precise modeling of mechanical components.
2D Parametric Primitive Analysis(PPA) aims to automate the understanding and analysis of 2D engineering drawings.
Vitruvion~\cite{seff2021vitruvion} leverages an autoregressive model to synthesize constraint-based sketches.
PICASSO~\cite{karadeniz2024picasso} introduces a feed-forward framework leveraging rendering self-supervision to infer parametric primitives from sketch images without explicit parameter annotations. 
Additionally, PpaCAD~\cite{wang2024parametric} formulates CAD sketch analysis as a set prediction task, effectively handling primitive extraction and constraint inference in a unified manner. More recently, DAVINCI~\cite{karadeniz2024davinci} proposes a single-stage architecture for constrained CAD sketch inference, streamlining the parameterization process by directly predicting both primitives and constraints.
However, existing PPA studies primarily focus on the geometric layer of engineering drawings while neglecting the annotation layer, limiting their applicability to real-world industrial scenarios.

\section{ParaCAD: Benchmarking 2D PPA}
As previously discussed, the absence of suitable datasets remains a critical barrier to advancing 2D PPA. Existing datasets primarily consist of sketches and hand-drawn images, which significantly differ from real-world engineering drawings in two fundamental aspects:
\begin{itemize}
    \item 
\textbf{Lack of Annotation Layer}: Engineering drawings encompass both a geometry layer, which encodes geometric primitives and constraints, and an annotation layer that provides essential manufacturing information, including dimensional annotations, functional symbols, and process instructions. In contrast, sketches and hand-drawn images predominantly capture only the geometry layer, omitting crucial parametric annotations required for precise design interpretation.
\item \textbf{Higher Structural Complexity and Real-World Constraints}: Unlike sketches and hand-drawn images, engineering drawings exhibit significantly greater structural complexity, incorporating a larger number of geometric primitives, intricate inter-relationships, and higher information density. Moreover, they must comply with real-world physical constraints to ensure manufacturability and functional validity (\textit{e.g.}, the requirement for fully enclosed shapes). Additionally, engineering drawings often contain interference elements, such as annotation lines, which introduce additional challenges for 2D PPA models.
\end{itemize}
\noindent To address the lack of PPA dataset, we introduce ParaCAD, the first large-scale benchmark that explicitly integrates the annotation layer and real-world industrial drawings, offering a more comprehensive and representative dataset for advancing 2D PPA research. Some samples of ParaCAD are shown in Fig.~\ref{sample}.

\begin{figure*}[t]
    \centering
    \includegraphics[width=1\linewidth]{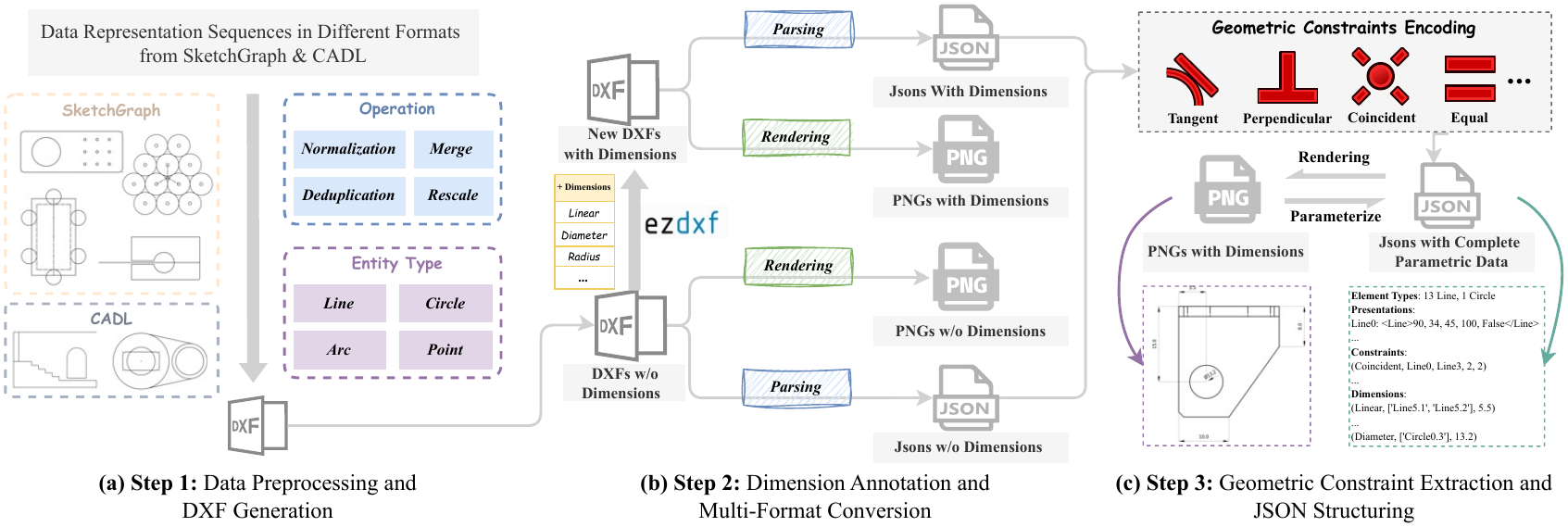}
\caption{
Figure: Data processing pipeline for engineering drawing parameterization. (a) Step 1: Preprocessing and DXF generation from SketchGraph and CADL datasets. (b) Step 2: Dimension annotation and conversion to multiple formats (DXF, PNG, JSON). (c) Step 3: Extraction of geometric constraints and structuring of parameterized data.
}
\label{fig:annotation_pipeline}
\end{figure*}

\subsection{Training Set with Annotation Layer}
ParaCAD consists of over 10 million annotated engineering drawings. The training set is divided into three subsets: single primitive recognition data, sketch structural perceiving data, and dimensional annotated drawing data, each tailored for different stages of model training. The first two subsets are derived from two large-scale open-source datasets, SketchGraphs\cite{seff2020sketchgraphs} and Computer-Aided Design as Language (CADL)\cite{ganin2021computer}. To enhance the effectiveness of the single primitive recognition data, we manually incorporated color-coded annotation lines, improving the model’s ability to distinguish and interpret individual geometric primitives.
For the construction of the dimensional annotated drawing data, we develop a systematic data processing pipeline, as illustrated in Figure~\ref{fig:annotation_pipeline}. The detailed processing steps are outlined as follows:

\textbf{Step 1: Data Preprocessing and DXF Generation}
First, we perform data preprocessing, where we decompose each sketch into four types of atomic components (\textit{i.e.}, \texttt{arcs}, \texttt{circles}, \texttt{lines}, and \texttt{points}) and a non-empty set of constraints. Then, we exclude any sketches with fewer than six primitives as they often consist of overly simplistic shapes with limited values (\textit{e.g.}, basic rectangles). To increase the complexity of dataset and ensure a diverse representation of geometric structures, we apply different filtering criteria to each dataset: For SketchGraphs, we follow~Vitruvion\cite{seff2021vitruvion} to retain sketches containing 6 to 30 primitives; for CADL, we reserve sketches with 1 to 25 primitives. Further, after deduplication and filtering, we obtain approximately 4 million unique sketches in total. Finally, we employ ezdxf\footnote{https://github.com/mozman/ezdxf} to transform reserved sketches into corresponding DXF files.

\textbf{Step 2: Dimension Annotation and Multi-Format Conversion}
In the following, we further process the generated DXF files, which only contain geometric primitives. Specifically, we use ezdxf to add dimensional parameter annotations to sketches, obtaining the new DXF files with dimensions. As a result, there are two versions of DXF files: non-annotated DXF files containing only geometric primitives, and annotated DXF files with dimension annotations.
Both of them are then converted into structured JSON files (for compatibility with VLM pipelines) and PNG images (providing a visual representation of 2D engineering drawings).

\textbf{Step 3: Geometric Constraint Extraction and JSON Structuring}
In the final step, we process the JSON files generated in Step 2 by analyzing positional relationships between primitives to extract geometric constraints. These constraints are then inserted into corresponding JSON files, ensuring that they contain three types of information:
\begin{itemize}
    \item \textit{Primitive Information}: fundamental geometric entities.
    \item \textit{Constraint Information}: relationships between primitives (\textit{e.g.}, parallelism and tangency).
    \item \textit{Dimension Annotation Information}: extracted dimension labels and numerical values.
\end{itemize}

\subsection{Evaluation sets with Real-World Industrial Drawings}
To better assess the performance of PPA models in real-world scenarios, we introduce a highly challenging evaluation dataset comprising 3,000 real-world 2D engineering drawings. Based on our data processing methodology, we obtain DXF files containing both the geometric layer and annotation layer as evaluation set.
Unlike open-source sketch datasets, these real-world drawings exhibit significantly higher complexity in terms of the number of geometric primitives, topological relationships, and visual information density. Additionally, real-world engineering drawings follow strict design intent and physical constraints, such as fully enclosed structures, functional dimensions, and standardized annotations.

\subsection{Evaluation Metric}
Existing 2D PPA evaluation metrics can be broadly categorized into two types: (1) Primitive-level metrics: Primitive F1 Score (\textbf{PF1})~\cite{karadeniz2024davinci}, Parametric Mean Squared Error (\textbf{ParamMSE})~\cite{karadeniz2024picasso}, Constraint F1 Score (\textbf{CF1})~\cite{karadeniz2024davinci}, and Accuracy (\textbf{Acc})~\cite{seff2021vitruvion,karadeniz2024davinci,wang2024parametric}. These metrics assess the alignment between predicted and ground-truth primitives and constraints. (2) Vision-based metrics: Image Mean Squared Error (\textbf{ImgMSE})~\cite{karadeniz2024picasso} and Chamfer Distance (\textbf{CD})~\cite{karadeniz2024picasso,karadeniz2024davinci}. They evaluate prediction quality in the image space.

Notably, existing 2D PPA methods do not consistently adopt all of these evaluation metrics, leading to fragmented benchmarking across different works. To address this, we reproduce and standardize existing methods while supplementing missing evaluation metrics, ensuring a comprehensive and unified assessment framework. In addition, existing evaluation metrics primarily assess the accuracy of predictions related to the geometry layer, overlooking the crucial role of annotation layer association in real-world industrial design. In practical applications, the success of PPA is not solely determined by the correct recognition of geometric primitives but by whether the geometry layer is correctly matched with the annotation layer. 

To quantitatively assess the accuracy of this association, we introduce a new metric Dimension Accuracy (\textbf{DA}), which evaluates the alignment between predicted dimensions and their corresponding ground-truth annotations. DA integrates three key validation functions: type correctness, numerical consistency, and geometric element alignment. It can ensure a comprehensive evaluation of how well dimensions are extracted and associated with their respective geometric elements. The formulation of DA is defined as follows:

\begin{equation}
T(P_i, \hat{P}_i) = \mathbb{I} \left( \operatorname{Type}(\hat{P}_i) = \operatorname{Type}(P_i) \right)
\label{eq:type_correctness}
\end{equation}

\begin{equation}
V(P_i, \hat{P}_i) = \mathbb{I} \left( \left| \hat{V}_i - V_i \right| \leq \tau_v \right)
\label{eq:numerical_consistency}
\end{equation}

\begin{equation}
E(P_i, \hat{P}_i) = \mathbb{I} \left( \sum_{k=1}^{N_i} \mathbb{I} \left( \left| \hat{E}_{i,k} - E_{i,k} \right| \leq \tau_e \right) = N_i \right)
\label{eq:geometric_element_alignment}
\end{equation}

\begin{equation}
DA = \frac{1}{T} \sum_{i=1}^{M} T(P_i, \hat{P}_i) \cdot V(P_i, \hat{P}_i) \cdot E(P_i, \hat{P}_i)
\label{eq:dimension_accuracy}
\end{equation}

Specifically, $P_i$ denotes the ground truth of primitives, and $\hat{P}_i$ represents the predicted result of primitives. The component \( T(P_i, \hat{P}_i) \) ensures that the predicted dimension type matches the ground truth, identifying whether it is length, diameter, radius, or angle.
The component \( V(P_i, \hat{P}_i) \) checks if the predicted dimension value deviates within \( \tau_v \).  
The component \( E(P_i, \hat{P}_i) \) ensures geometric elements align within the positional tolerance \( \tau_e \).
Dimension Accuracy (DA) measures the ratio of correctly predicted dimensions to total ground truth dimensions, requiring all three conditions—type correctness, numerical consistency, and geometric element alignment—to be met.

 \subsection{Evaluation Paradigm}
We evaluate both existing PPA methods and the proposed PHT-CAD using three distinct evaluation paradigms: (1) \textit{Standard Evaluation}, where the model is trained on each dataset’s training set and evaluated on its corresponding test set to simulate the standard PPA process; (2) \textit{Zero-shot Evaluation}, in which the model is trained on one dataset’s training set and evaluated on unseen datasets’ test sets to simulate a zero-shot scenario; and (3) \textit{Dimension-based Evaluation}, where we assess the accuracy of the matching between the geometry and annotation layers using the proposed DA metric.

\begin{figure*}[t]
    \centering
    \includegraphics[width=0.9\textwidth]{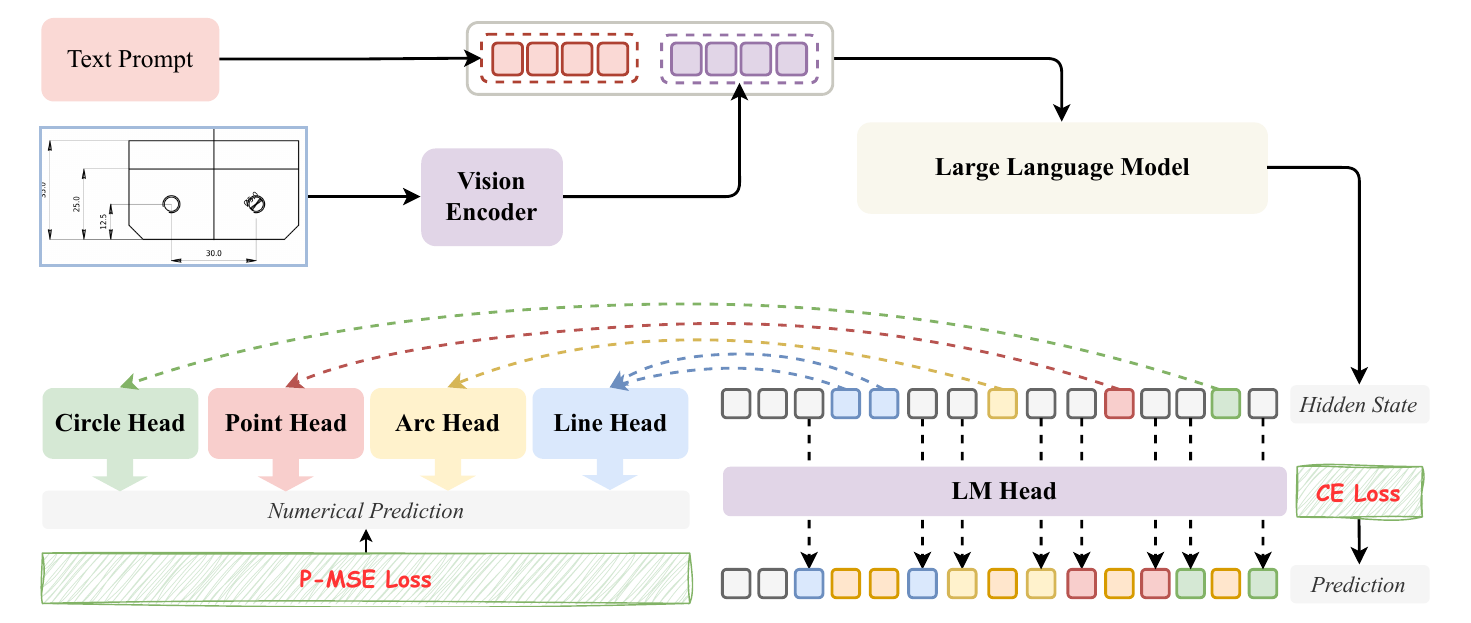} 
    \caption{Overview of PHT-CAD framework. }
    \label{framework}
\end{figure*}
\section{Methodology}
In this paper, we introduce an innovative 2D PPA method called PHT-CAD, which enables precise engineering drawing analysis. The subsequent sections present the proposed Efficient Hybrid Parametrization (EHP) representation for 2D engineering drawings, the PHT-CAD framework, and the Progressive Hierarchical Tuning (PHT) strategy.

\subsection{Efficient Hybrid Parametrization} 
Existing parametrization strategies can be broadly classified into three main types: (1) \textbf{Implicit Strategy} utilizes a normalized and relative representation of geometric primitives, encoding their spatial properties through direction vectors, reference points, and parametric constraints. (2) \textbf{Point-Based Strategy} adopts a normalized absolute representation, encoding geometric primitives via explicit key points rather than relative constraints. (3) \textbf{Overparameterized Strategy} combines parameters from both the Implicit and Point-based strategies, aiming to enrich the model's geometric information by incorporating both relative constraints and explicit key points.

Building upon these existing parametrization strategies, we propose a new representation Efficient Hybrid Parametrization (EHP) for 2D engineering drawings. EHP integrates the point-based and implicit strategies while eliminating redundant information to enhance efficiency and consistency. EHP introduces three key modifications: (1) EHP removes direction vectors and instead represents lines and arcs solely through their start and end coordinates, allowing the direction to be inferred naturally from these points. (2) EHP redefines the representation of circles and arcs using their center coordinates, radius, and start/end angles, moving away from discrete point-based representations. This formulation enables the model to directly infer constraint relationships and dimensional attributes, while reinforcing its ability to learn the topological dependencies between the center and radius. (3) EHP normalizes coordinates to the range [0, 1000), providing a relative coordinate system that ensures consistent spatial scaling across images of varying resolutions. This normalization facilitates the model's learning of positional relationships and improves its robustness to variations in input dimensions. In summary, EHP can be described as follows:

\[
\begin{array}{ll}
\textbf{Point:} & p = (x_p, y_p) \\
\textbf{Line:} & l = (x_\text{start}, y_\text{start}, x_\text{end}, y_\text{end}, v) \\
\textbf{Circle:} & c = (x_c, y_c, r) \\
\textbf{Arc:} & a = (x_a, y_a, r, \theta_\text{start}, \theta_\text{end})
\end{array}
\]
where \( (x_{start}, y_{start}) \) and \( (x_{end}, y_{end}) \) denote the start and end coordinates, and \( v \) is a binary indicator specifying validity (\textit{e.g.}, solid or dashed). \( (x_c, y_c) \) is the center coordinate and \( r \) is the radius. \( \theta_\text{start}, \theta_\text{end} \) are the start and end angles, respectively.

\begin{figure*}[t]
    \centering
    \includegraphics[width=1\textwidth]{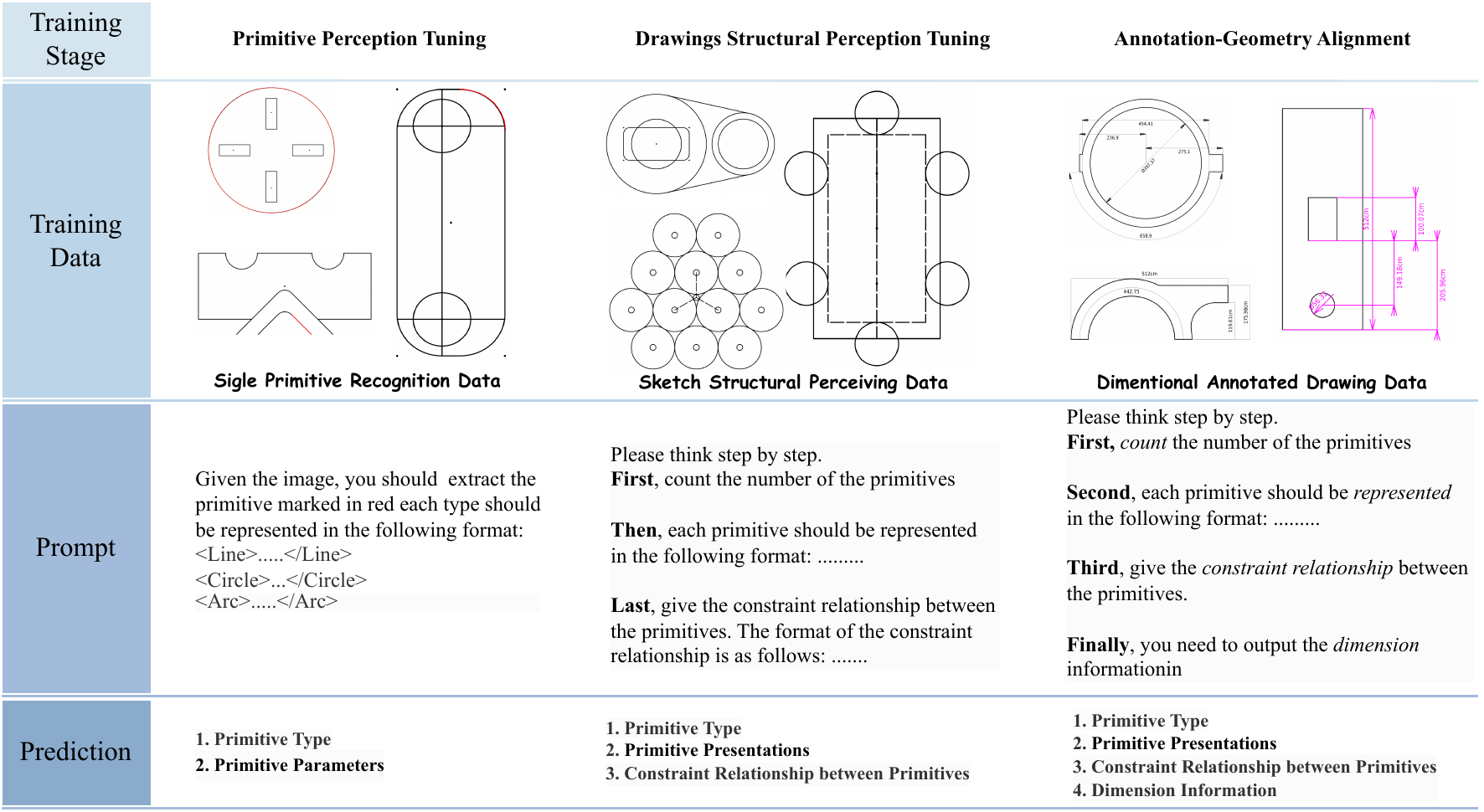} 
    \caption{Overview of the Progressive Hierarchical Tuning.}
    \label{PHT}
\end{figure*}

\subsection{Framework}
The detailed architecture of PHT-CAD is shown in Fig.~\ref{framework}. The PHT-CAD framework integrates a ViT-based visual encoder with a text encoder derived from Qwen2.5~\cite{yang2024qwen2}, enabling effective multi-modal understanding of 2D engineering drawings. The model is trained by the proposed Progressive Hierarchical Tuning (PHT) strategy, which consists of a stepwise process that progressively enhances PHT-CAD's capability for structured CAD parsing and fine-grained parametric primitive analysis.

\noindent \textbf{Vision Encoder.} The vision encoder in PHT-CAD is built upon the Vision Transformer (ViT)~\cite{dosovitskiy2020image}, serving as a robust feature extractor for analyzing the geometric structures of CAD drawings. This encoder captures both global and local features, ensuring a comprehensive understanding of the CAD elements. Specifically, it extracts global structural information to capture the overall layout and spatial relationships between primitives, while simultaneously focusing on local geometric details for accurate representation of individual primitives.

\noindent \textbf{Text Decoder.} The text decoder in PHT-CAD is based on Qwen2.5, which decodes the visual feature embeddings and translates them into structured CAD representations. Building on the proposed EHP, the text decoder outputs a structured parametric representation of the input 2D engineering drawing. To enhance the model’s ability to refine geometric parameterization, we introduce special tokens corresponding to four types of atomic components: circle, point, arc, and line. These tokens enable precise localization of the hidden states associated with these primitives. Once the hidden states are retrieved, they are passed as inputs to four dedicated regression heads. Each head is tailored to predict the parametric attributes of corresponding geometric primitives, thereby ensuring accurate numerical estimations.

\noindent \textbf{Loss Function.}
Existing Vision-language Models (VLMs) typically employ Cross-Entropy (CE) loss for optimization. While CE loss is effective for maximizing the likelihood of categorical labels, it does not explicitly account for numerical discrepancies between predicted and ground-truth values, making it unsuitable for fine-grained parameter estimation in the 2D PPA task. In contrast, Mean Squared Error (MSE) loss penalizes deviations quadratically, ensuring that even small numerical discrepancies in geometric parameters are captured and optimized effectively. This characteristic is particularly crucial for PPA since maintaining high numerical precision is essential to preserve the structural integrity of CAD applications.
To this end, we introduce a novel Parametric Mean Squared Error (P-MSE) loss, which enhances the precision of primitive parameterization. The P-MSE loss supervises the numerical predictions generated by the four dedicated regression heads, enforcing a fine-grained alignment between predicted and ground-truth parametric values. Thus, the CE loss $\mathcal{L}_{\text{CE}}$ is employed to supervise the output of the original language model head, while the P-MSE loss $\mathcal{L}_{\text{P-MSE}}$ is computed between the numerical predictions from the four dedicated regression heads and their corresponding ground-truth values. The total loss $\mathcal{L}$ is then obtained by a weighted summation of $\mathcal{L}_{\text{CE}}$ and $\mathcal{L}_{\text{P-MSE}}$:

\begin{equation}
\mathcal{L}_{\text{CE}} = - \sum_{i=1}^{N} t_i \log \hat{t}_i
\end{equation}

\begin{equation}
\mathcal{L}_{\text{P-MSE}} = \frac{1}{N} \sum_{i=1}^{N} \left| f_{\theta_i} (h_i) - p_i \right|^2
\end{equation}

\begin{equation}
\mathcal{L} = \lambda_{\text{CE}} \mathcal{L}_{\text{CE}} + \lambda_{\text{P-MSE}} \mathcal{L}_{\text{P-MSE}}
\end{equation}
where \( t_i \) represents the ground truth and \( \hat{t}_i \) represents the predicted probability distribution. \( f_{\theta_i} (h_i) \) denotes the MLP-based regression head applied to the hidden representation \( h_i \) extracted from the special token corresponding to each primitive, and \( p_i \) represents the ground truth parameter. $N$ represents the number of predicted tokens. \( \lambda_{\texttt{CE}} \) and \( \lambda_{\texttt{P-MSE}} \) are two hyperparameters to balance classification and regression objectives.

\subsection{Progressive Hierarchical Tuning}
The proposed Progressive Hierarchical Tuning (PHT) strategy consists of three stages (shown in Fig.~\ref{PHT}):

\begin{table*}[t]
\centering
\small
\setlength{\tabcolsep}{3pt} 
\begin{tabular}{l@{\hskip 3pt}cccccc@{\hskip 8pt}cccccc@{\hskip 8pt}cccccc}
\toprule
\multirow{3}{*}{\textbf{\textsc{Methods}}} & \multicolumn{12}{c}{\textbf{SketchGraph}}  \\ 
\cmidrule(lr){2-13} 
& \multicolumn{6}{c}{Precise Sketch Images} & \multicolumn{6}{c}{Hand-drawn Sketch Images}  \\ 
\cmidrule(lr){2-7} \cmidrule(lr){8-13} 
& Acc$\uparrow$ & ParamMSE$\downarrow$ & ImgMSE$\downarrow$ & CD$\downarrow$ & CF1$\uparrow$ & PF1$\uparrow$
& Acc$\uparrow$ & ParamMSE$\downarrow$ & ImgMSE$\downarrow$ & CD$\downarrow$ & CF1$\uparrow$ & PF1$\uparrow$
 \\ 
\midrule
ResNet34~\cite{he2016deep}    & 0.465 & 908  & 0.199 & 5.883 & --- & ---  & 0.396 & 1048 & 0.240 & 6.908 & --- & ---  \\ 
PpaCAD~\cite{wang2024parametric}  & 0.524 & 589  & 0.195 & 5.097 & --- & ---  & 0.464 & 744  & 0.244 & 6.904 & --- & ---   \\ 
Vitruvion~\cite{seff2021vitruvion} & 0.537 & 624  & 0.186 & 4.901 & 0.238 & 0.706  & 0.461 & 685  & 0.237 & 5.258 & --- & --- \\ 
PICASSO~\cite{karadeniz2024picasso}  & 0.751 & 281  & 0.075 & 0.729 & --- & ---  & 0.658 & 365  & 0.117 & 1.090 & --- & ---   \\ 
\midrule
PHT-CAD (w/o hand) & \textbf{0.859} & \textbf{52}  & \textbf{0.003} & --- & \textbf{0.868} & \textbf{0.917}  & ---  & --- & --- & --- & --- & ---  \\ 
PHT-CAD & 0.811 & {55}  & {0.004} & \textbf{0.008} & 0.784 & 0.879  & \textbf{0.795}  & \textbf{11}  & \textbf{0.005} & \textbf{0.010} & \textbf{0.7618} & \textbf{0.8665}   \\ 
\bottomrule
\end{tabular}
\caption{Comparison of PHT-CAD with SOTA methods in standard evaluation paradigm. ‘(w/o hand)’ indicates that hand-drawn sketch images are not included in the training data.}
\label{standard}
\end{table*}

\begin{table}[t]
\centering
\small
\setlength{\tabcolsep}{1pt} 
\begin{tabular}{l@{\hskip 3pt}cccccc@{\hskip 8pt}cccccc@{\hskip 8pt}cccccc}
\toprule
\textbf{\textsc{Methods}}

& Acc$\uparrow$ & ParamMSE$\downarrow$ & ImgMSE$\downarrow$ & CD$\downarrow$ & CF1$\uparrow$ & PF1$\uparrow$ \\ 
\midrule
ResNet34~\cite{he2016deep}     & 0.520 & 829  & 0.189 & 5.698 & --- & --- \\ 
PpaCAD~\cite{wang2024parametric}   & 0.562 & 601 & 0.272 & 6.600 & --- & --- \\ 
Vitruvion~\cite{seff2021vitruvion}   & 0.560 & 608  & 0.190 & 5.568 & 0.242 & 0.710 \\ 
PICASSO~\cite{karadeniz2024picasso}  & 0.809 & 199  & 0.067 & 0.739 & --- & --- \\ 
\midrule
PHT-CAD   & \textbf{0.923} & \textbf{50}  & \textbf{0.003} & \textbf{0.106} & \textbf{0.860} & \textbf{0.910} \\ 
 
\bottomrule
\end{tabular}
\caption{Comparison of PHT-CAD with SOTA methods in zero-shot evaluation paradigm.}
\label{zero-shot}
\end{table}

\begin{table}[t]
\centering
% \small
\setlength{\tabcolsep}{1pt} 
\begin{tabular}{l@{\hskip 3pt}cccccc@{\hskip 8pt}cccccc@{\hskip 8pt}cccccc}
\toprule
\textbf{\textsc{Methods}}

& Acc$\uparrow$ & ParamMSE$\downarrow$ & ImgMSE$\downarrow$ & CD$\downarrow$ & CF1$\uparrow$ & PF1$\uparrow$ \\ 
\midrule
PHT-CAD   & 0.840 & 50  & 0.002 & --- & 0.860 & 0.910 \\ 
 
\bottomrule
\end{tabular}
\caption{The performance of PHT-CAD under dimension-based evaluation.}
\label{da}
\end{table}

1) \textbf{Stage 1: Primitive Perception Tuning.} In this stage, the model is trained to recognize and classify individual geometric primitives, outputting their parameters in a structured, parametric format. Primitive Perception Tuning provides a foundational understanding of basic geometric structures, which is essential for more complex tasks in subsequent stages. During this stage, we utilize the \textit{single primitive recognition} data of ParaCAD. We initialize PHT-CAD with the 0.5B pretrained weights from GOT~\cite{wei2024general}, leveraging GOT's capabilities in geometric shape recognition, structured text parsing, and spatial relationship modeling. 

2) \textbf{Stage 2: Structural Perception Tuning.} In this stage, we expand the model’s ability to perceive the full range of primitives in engineering drawings, and to understand the inter-dependencies and constraints among them. During this stage, we use the \textit{sketch structural perceiving} data of ParaCAD. The training process instructs the model to first count the number of primitives, followed by outputting the parameters of each primitive according to the proposed EHP, and then analyzing the geometric constraint relationships between them. This training stage enables the model to learn the geometric structure in the image in an end-to-end manner and generate outputs in a standardized, structured format. The model is initialized with the parameters trained in the first stage.

3) \textbf{Stage 3: Annotation-geometry Alignment.} The final tuning stage aims to enhance the model’s ability to process engineering drawings that include dimensional annotations, while simultaneously predicting the primitives, constraints, and dimensional information. In this stage, we utilize the \textit{dimensional annotated drawing} data of ParaCAD.

Through the PHT strategy, the model's intrinsic geometric perception and reasoning abilities are progressively enhanced, allowing for fine-grained primitive parameterization. All parameters of PHT-CAD are optimized across three sequential tuning stages.

\section{Experiment}

\begin{table*}[t]
\centering
\small
\setlength{\tabcolsep}{3pt} 
\begin{tabular}{l@{\hskip 3pt}cccc@{\hskip 8pt}ccccc@{\hskip 8pt}cccc@{\hskip 8pt}ccccc}
\toprule
\multirow{2}{*}{\textbf{\textsc{Methods}}} & \multicolumn{5}{c}{\textbf{SketchGraph}} & \multicolumn{5}{c}{\textbf{ParaCAD}} \\ 
\cmidrule(lr){2-6} \cmidrule(lr){7-11}

& Acc$\uparrow$ & ParamMSE$\downarrow$ & CF1$\uparrow$ & PF1$\uparrow$ & DA$\uparrow$

& Acc$\uparrow$ & ParamMSE$\downarrow$ & CF1$\uparrow$ & PF1$\uparrow$ & DA$\uparrow$ \\ 
\midrule

Stage 1 + 3      
                & 0.72  & 64.44  &  0.66   & 0.80   & 41.31/80.80  
               
                & 0.49  & 48.42  &  0.42   & 0.63   & 0.30/0.49 \\ 
Stage 2 + 3      
                & 0.75  & 64.44  &  0.70   & 0.82   & 45.31/82.80  
               
                & 0.54  & 46.42  &  0.58   & 0.76   & 0.33/0.51 \\ 
Stage 1 + 2 + 3 
                & 0.87  & 55.01  & 0.90 & 0.92 & 60.13/88.29
                  
                & 0.76  & 42.11  & 0.79 & 0.85 & 0.47/0.69 \\ 
\bottomrule
\end{tabular}
\caption{Comparison of different training stages on various evaluation metrics for SketchGraph and CADL datasets, including non-dimensioned and dimensioned data with the DA metric.}
\label{stage}
\end{table*}

\begin{table*}[t]
\centering

\setlength{\tabcolsep}{3pt} 
\begin{tabular}{l@{\hskip 3pt}cccccc@{\hskip 8pt}ccccc}
\toprule
\multirow{2}{*}{\textbf{\textsc{Methods}}} & \multicolumn{5}{c}{\textbf{SketchGraph}} & \multicolumn{5}{c}{\textbf{ParaCAD}} \\ 
\cmidrule(lr){2-6} \cmidrule(lr){7-11}
& Acc$\uparrow$ & ParamMSE$\downarrow$ & CF1$\uparrow$ & PF1$\uparrow$ & ImgMSE$\downarrow$  
& Acc$\uparrow$ & ParamMS$\downarrow$E & CF1$\uparrow$ & PF1$\uparrow$ & ImgMSE$\downarrow$ \\ 
\midrule
w/o P-MSE Loss & 0.79 & 56.27  & 0.77 & 0.85 & 0.0065  
             & 0.71 & 40.40  & 0.69 & 0.75 & 0.01 \\ 
w/ P-MSE Loss  & 0.86 & 51.61  & 0.87 & 0.92 & 0.0027  
             & 0.75 & 39.63  & 0.78 & 0.84 & 0.0081 \\ 

\bottomrule
\end{tabular}
\caption{Comparison of methods with and without P-MSE Loss on various evaluation metrics for SketchGraph and ParaCAD datasets.}
\label{mse_loss_comparison}
\end{table*}

\subsection{Implementation Details} 
Our model is trained with 8 NVIDIA A100 GPUs. In the first stage, we optimize all parameters with a batch size of 8 for 2 epochs, where we use the AdamW optimizer with a cosine annealing scheduler, an initial learning rate of 1e-4, and a maximum token length of 4096. In the second stage, we expand the maximum token length to 8192 while training the model with the same optimization setting for 1 epoch. Finally, in the third stage, we introduce dimension annotations while ensuring the model retain its generalization ability for sketches without dimensions. The learning rate is set to 2e-5, and 50\% data of the second stage is sampled to maintain the model’s performance on non-dimensioned sketches.

\subsection{Evaluation Results}

\noindent \textbf{Standard Evaluation.} 
As shown in Tab.~\ref{standard}, we present the results of PHT-CAD in standard evaluation paradigm on the SketchGraph dataset. PHT-CAD demonstrates outstanding performance, achieving a 6\% improvement in accuracy on Precise Sketch Images compared to SOTA method. 
The improvement is more pronounced when the domain gap between the training and test data is smaller (i.e., when hand-drawn data is not included), achieving a 10.8\% increase.
On Hand-drawn Sketch Images, PHT-CAD exhibits a 13.7\% accuracy boost. These significant improvements are primarily attributed to the enhancements introduced in Stage 1 and Stage 2 of the training process, which improve the model’s capability to perceive primitives and learn structural constraints.

\noindent \textbf{Zero-shot Evaluation.} 
In Tab.~\ref{zero-shot}, we report the results of PHT-CAD compared to SOTA methods in a zero-shot evaluation paradigm. We take SketchGraph as training set and CADL as test set. PHT-CAD outperforms all other methods across all metrics, achieving an 11.4\% improvement in accuracy. This notable performance gain is largely attributed to the model's ability to leverage the structural constraint reasoning and semantic understanding capabilities inherent in VLMs, granting the model strong generalization capabilities.

\noindent \textbf{Dimension-based Evaluation.}
We first introduce ParaCAD, a benchmark designed for Dimension-based Evaluation, and evaluate the performance of PHT-CAD under this setting. Due to the lack of prior work in this area, we only report the performance of PHT-CAD on ParaCAD dataset to assist the community in further exploring related research. As shown in Table~\ref{da}, the proposed PHT-CAD achieves 84.0\% in the metric of Acc.

\begin{figure*}[t]
    \centering
    \includegraphics[width=0.9\textwidth]{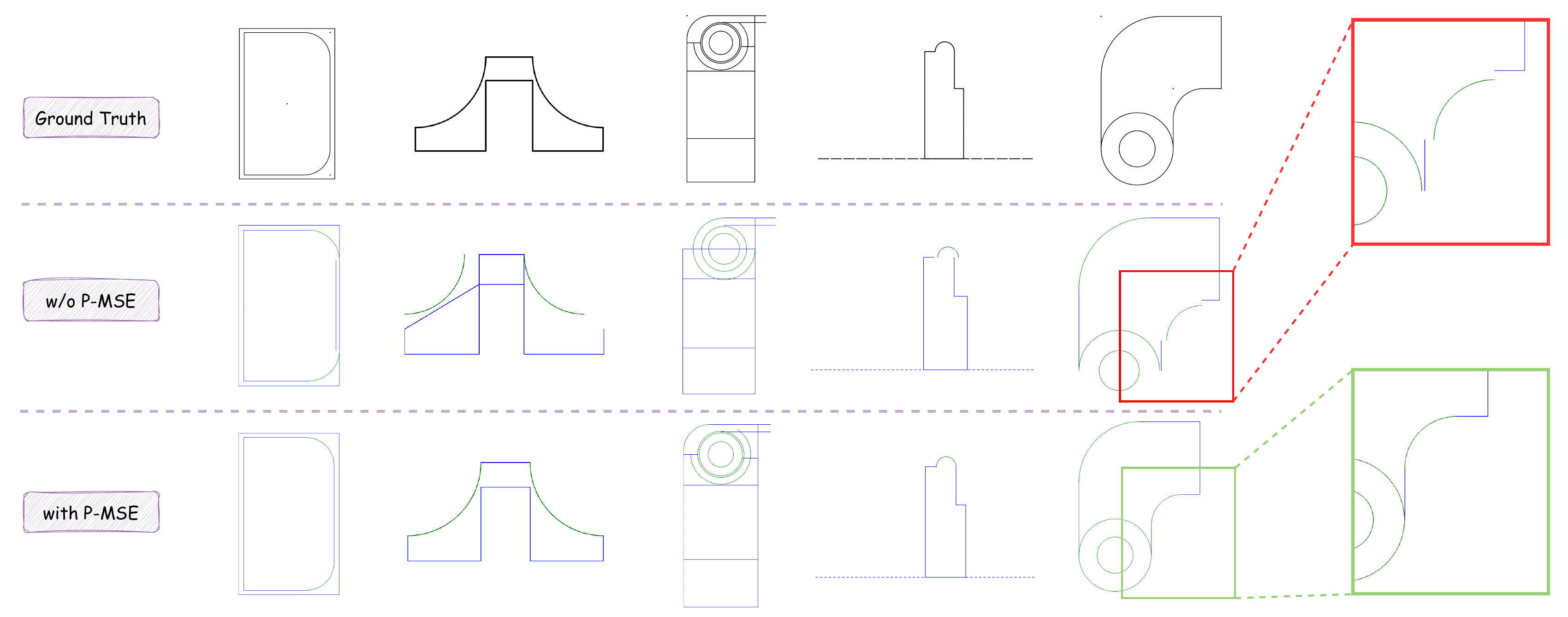} 
    \caption{Visual comparison of CAD drawing reconstruction performance with and without P-MSE loss. }
    \label{pmse}
\end{figure*}

\subsection{Abalation Study}
\textbf{Three-Stage PHT Strategy.} 
We conduct ablation experiments on Stage 1 and Stage 2 separately. The results are shown in Tab.~\ref{stage}. The performance dropped significantly in the absence of Stage 1, with accuracy decreasing by 12\%. This decline can be attributed to the fact that Stage 1 is trained to recognize and classify individual geometric primitives, which lays a critical foundation for the model to perceive the entire drawing. Compared to the absence of Stage 1, the performance degradation is even more pronounced when Stage 2 is unused. Specifically, the performance of PHT-CAD will decrease by 15\% in accuracy, which further demonstrates the crucial importance of perceiving the full range of primitives in engineering drawings.

\noindent \textbf{P-MSE loss.} As shown in Tab.~\ref{mse_loss_comparison}, we compare the performance of models with and without P-MSE Loss on the SketchGraph and ParaCAD datasets. The results show that the model with P-MSE Loss achieves significant performance improvements on both datasets. P-MSE Loss effectively enhances the model's accuracy and image reconstruction quality, with more pronounced effects on the SketchGraph dataset. Further, as shown in Fig.~\ref{pmse}, we provide a visual comparison between the models with and without P-MSE. It is evident that after incorporating P-MSE, the model demonstrates improved perception of fine-grained details of primitives.

\section{Conclusion}
In this paper, we introduce PHT-CAD, an innovative framework for 2D Parametric Primitive Analysis (PPA) that harnesses the power of Vision-Language Models (VLMs) for precise engineering drawing analysis. By proposing the Efficient Hybrid Parametrization (EHP) strategy, we enhance the precision and consistency of parametric representations, demonstrating its effectiveness through extensive experiments. Additionally, we present ParaCAD, the first large-scale 2D PPA benchmark that incorporates both the geometric and annotation layers, addressing the existing data gap and improving real-world applicability. Our new evaluation metric, Dimension-based Evaluation, assesses the matching accuracy between the geometry and annotation layers. Experimental results across various benchmarks validate the effectiveness of PHT-CAD, marking a significant advancement in 2D PPA for industrial design and manufacturing applications.

{
    \small
    \bibliographystyle{ieeenat_fullname}
    \bibliography{main}
}

\end{document}